%% file: Template_Regular.tex
\documentclass{article}
\usepackage{spconf,amsmath,graphicx}
\usepackage{spconf,amsmath,graphicx}
\usepackage{cite}
\usepackage{amssymb}
\usepackage{hyperref}
\usepackage{soul}

\usepackage{placeins}
\usepackage{array}
\usepackage{booktabs,colortbl}
\usepackage{tabularx}
\usepackage{multirow}

\newcommand{\PreserveBackslash}[1]{\let\temp=\\#1\let\\=\temp}
\newcolumntype{C}{>{\centering\arraybackslash}X}
\newcolumntype{L}{>{\raggedright\arraybackslash}X}
\newcolumntype{R}{>{\raggedleft\arraybackslash}X}
\newcolumntype{A}[1]{>{\PreserveBackslash\raggedright}p{#1}}

\usepackage{acro}
\input{defined_acronyms}

\newcommand{\customDN}{{\small \scshape DAVEnet}}
\newcommand{\contrastiveDN}{{\small \scshape ContrastiveDAVEnet}}
\newcommand{\ma}{{\small \scshape LocalisationAttentionNet}}
\newcommand{\smallcustomDN}{{\footnotesize \scshape DAVEnet}}
\newcommand{\smallcontrastiveDN}{{\footnotesize \scshape ContrastiveDAVEnet}}
\newcommand{\smallma}{{\footnotesize \scshape LocalisationAttentionNet}}

\usepackage[prependcaption,textsize=scriptsize]{todonotes}
\definecolor{mycolor}{HTML}{FF6600}

\title{Towards visually prompted keyword localisation for zero-resource spoken languages}
\name{Leanne Nortje and Herman Kamper\thanks{Leanne Nortje is funded through a DeepMind PhD scholarship.}}
\address{MediaLab, Electrical \& Electronic Engineering, Stellenbosch University, South Africa}
\copyrightnotice{978-1-6654-7189-3/22/\$31.00~\copyright2023 IEEE}
\begin{document}
	%
	\maketitle
	\begin{abstract}
		Imagine being able to show a system a visual depiction of a keyword and finding spoken utterances that contain this keyword from a zero-resource speech corpus.
		We formalise this task and call it visually prompted keyword localisation (VPKL): given an image of a keyword, detect and predict where in an utterance the keyword occurs.
		To do VPKL, we propose a speech-vision model with a novel localising attention mechanism which we train with a new keyword sampling scheme.
		We show that these innovations give improvements in VPKL over an existing speech-vision model.
		We also compare to a visual bag-of-words (BoW) model where images are automatically tagged with visual labels and paired with unlabelled speech.
		Although this visual BoW can be queried directly with a written keyword (while our's takes image queries), our new model still outperforms the visual BoW in both detection and localisation, giving a 16\% relative improvement in localisation F1.
	\end{abstract}
	\acresetall
	\input{introduction}
	\input{kws}
	\input{multimodalModel}
	\input{experimentalSetup}
	\input{results}
	\acresetall
	\input{conclusions}
	
	\bibliographystyle{IEEEtran}
	{\bibliography{refs}}    
	
\end{document}

%% file: defined_acronyms.tex
\DeclareAcronym{davenet}{
	short = DAVEnet ,
	long  = deep audio-visual embedding network,
}
\DeclareAcronym{resdavenet}{
	short = ResDAVEnet,
	long  = residual deep audio-visual embedding network,
}
\DeclareAcronym{AE}{
	short = AE ,
	long  = autoencoder
}
\DeclareAcronym{CAE}{
	short = CAE,
	long  = correspondence autoencoder,
}
\DeclareAcronym{MCAE}{
	short = MCAE,
	long  = multimodal correspondence autoencoder,
}
\DeclareAcronym{MTriplet}{
	short = MTriplet,
	long  = multimodal triplet network,
}
\DeclareAcronym{FFNN}{
	short = FFNN,
	long  = feedforward neural network,
}
\DeclareAcronym{CNN}{
	short = CNN,
	long  = convolutional neural network,
}
\DeclareAcronym{RNN}{
	short = RNN,
	long  = recurrent neural network,
}
\DeclareAcronym{VQ}{
	short = VQ,
	long  = vector-quantisation
}
\DeclareAcronym{mfcc}{
	short = MFCC,
	long  = mel-frequency cepstral coefficient
}
\DeclareAcronym{CPC}{
	short = CPC,
	long  = contrastive predictive coding,
}
\DeclareAcronym{multdavenet}{
	short = MultDAVEnet,
	long  = multilingual deep audio-visual embedding network,
}
\DeclareAcronym{accmultdavenet}{
	short = AcstMultDAVEnet,
	long  = acoustic multilingual deep audio-visual embedding network,
}
\DeclareAcronym{DNN}{
	short = DNN,
	long  = deep neural network
}
\DeclareAcronym{ASR}{
	short = ASR,
	long  = automatic speech recognition
}
\DeclareAcronym{AWE}{
	short = AWE,
	long  = acoustic word embedding,
}
\DeclareAcronym{AGWE}{
	short = AGWE,
	long  = acoustically grounded word embedding,
}
\DeclareAcronym{ASE}{
	short = ASE,
	long  = acoustic span embedding,
}
\DeclareAcronym{QbE}{
	short = QbE,
	long  = query-by-example,
}
\DeclareAcronym{STD}{
	short = STD,
	long  = spoken term detection
}
\DeclareAcronym{UTD}{
	short = UTD, 
	long = long short-term memory
}
\DeclareAcronym{SGD}{
	short = SGD, 
	long = stochastic gradient decent,
}
\DeclareAcronym{bow}{
	short = BoW, 
	long = bag-of-words,
}
\DeclareAcronym{mse}{
	short = MSE, 
	long = mean squared error,
}
\DeclareAcronym{vpkl}{
	short = VPKL, 
	long = visually prompted keyword localisation,
}

%% file: introduction.tex
\section{Introduction}
\label{sec:intro}

How can we search a speech collection in a zero-resource language where it is impossible to obtain text transcriptions (e.g. unwritten languages)?
One way in which recent 
research is addressing 
this problem is 
to use vision as a weak form of supervision: speech systems are built on images paired with unlabelled spoken captions---removing the need for text~\cite{chrupala_visually_2022}.

\begin{figure}[!b]
	
	\begin{minipage}[b]{1.0\linewidth}
		\centering
		\centerline{\includegraphics[width=0.99\linewidth]{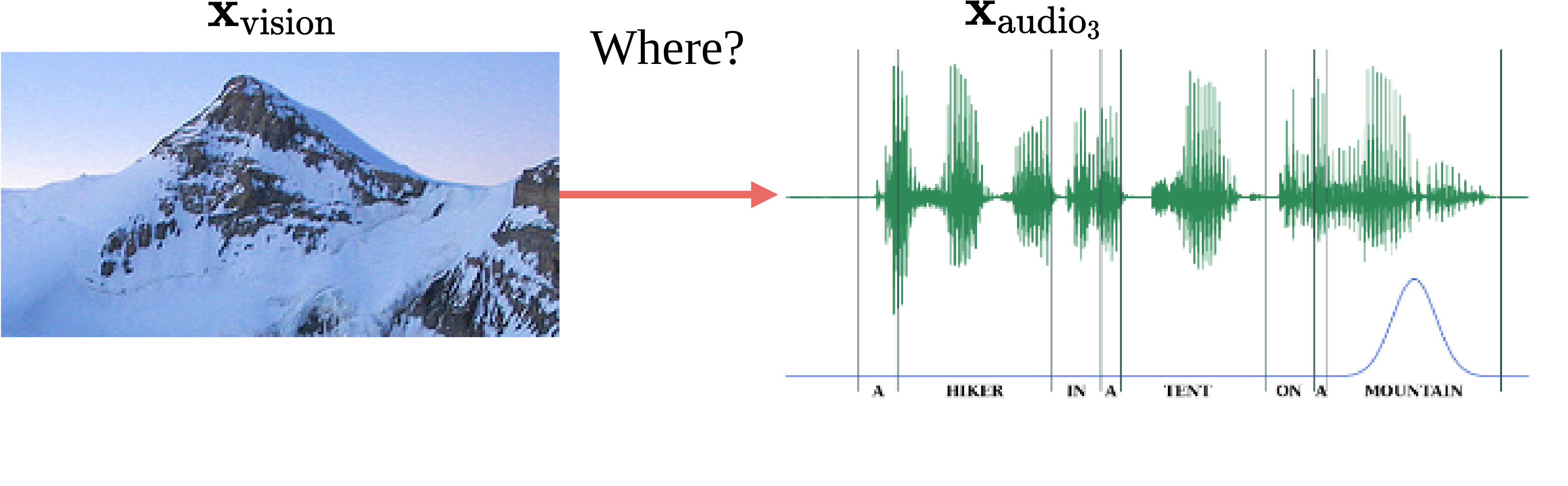}}
		\vspace{-25pt}
	\end{minipage}
	\caption{The goal in 
		visually prompted keyword localisation is to locate a given query keyword (given as an image) within a spoken utterance.}
	\label{fig:vpkl}
\end{figure}

In this paper we specifically introduce the new \ac{vpkl} task: a model is given an image depicting a keyword---serving as an image query---and is prompted to detect whether the query 
occurs in a spoken utterance.
If the keyword is detected, the model 
should also determine where in the utterance the keyword occurs.
E.g.\ the model is shown an image of a \textit{mountain} and asked whether it occurs in the 
spoken caption: ``a hiker in a tent on a mountain''. 
The model should also say where in the utterance
\textit{mountain} occurs (if it is detected),
as shown in Fig.~\ref{fig:vpkl}. 
To do this, we need a multimodal model that can compare images or image regions to
spoken utterances.

In the last few years a range of speech-vision models have been proposed~\cite{harwath_jointly_2018, harwath_learning_2017, harwath_unsupervised_2016, harwath_vision_2018, peng_word_2022, peng_fast-slow_2022, pasad_contributions_2019, chrupala_visually_2022}. Most were developed for retrieving whole images given a whole spoken caption as query (or vice versa).
Image-caption retrieval is different from VPKL---in the latter, the query is typically a depiction of an isolated object or concept and we want to detect and localise this query \textit{within} an utterance (rather than retrieving a whole spoken caption).
Nevertheless, with slight modification, we can use an image-caption retrieval model for VPKL.
We show that this performs poorly, 
presumably because of the mismatch between the training objective and the test-time VPKL task.

As a result, we propose 
a novel localising attention mechanism and a new keyword sampling scheme.
First, for the attention mechanism, we combine the idea of \textit{matchmaps}~\cite{harwath_jointly_2018} with a more explicit form of within-utterance attention~\cite{olaleye_attention-based_2021, tamer_keyword_2020}.
Second, for the sampling scheme, we can use a visual tagger to automatically tag training images with text labels of words likely occurring in the image.
From these generated tags, 
we can sample positive and negative image-caption pairs which contain the same or different keywords.
E.g.\ while originally we could have a spoken caption ``hikers going up a mountain slope'' paired only with a single image, we could now also pair this utterance with the 
spoken
caption ``a boy and his dad on a mountain.'' 
This would encourage the model to not only focus on utterances as a whole, but also learn within-utterance distinctions between keywords.
Note that in this paper we mainly consider an
idealised case using 
the captions' text 
transcriptions to sample positive and negative pairs
(simulating an ideal visual tagger). 

In this setting, we show that both innovations lead to improvements in VPKL over an image-caption retrieval model.  
We also compare to a visual \ac{bow} model~\cite{olaleye_towards_2020, olaleye_attention-based_2021, olaleye_keyword_2022}
which is queried with written keywords instead of images.
This model is trained using a visual tagger to
generate textual \ac{bow} labels for training 
images.
These labels are then used to train a keyword detection model~\cite{olaleye_keyword_2022}. 
While a written keyword arguably gives a stronger query signal than an image, we show that when combining our new attention mechanism with our new sampling scheme, we outperform the visual \ac{bow}.
Further analysis shows 
that the distribution of keywords that our model 
is able to localise is much smaller than that of the visual \ac{bow} model. 
We attribute this 
to image queries sometimes depicting more than one keyword.
Through further analyses, we also show that the model's performance decreases when tasked with learning a larger set of keywords. 
We also present initial experiments where a real image tagger is used to produce positive and negative examples for our training scheme---highlighting additional challenges for future work.

%% file: kws.tex
\section{New task: Visually prompted keyword localisation}
\label{sec:kws}

\begin{figure}[!b]
	
	\begin{minipage}[b]{1.0\linewidth}
		\centering
		\vspace{-5pt}
		\centerline{\includegraphics[width=1.0\linewidth]{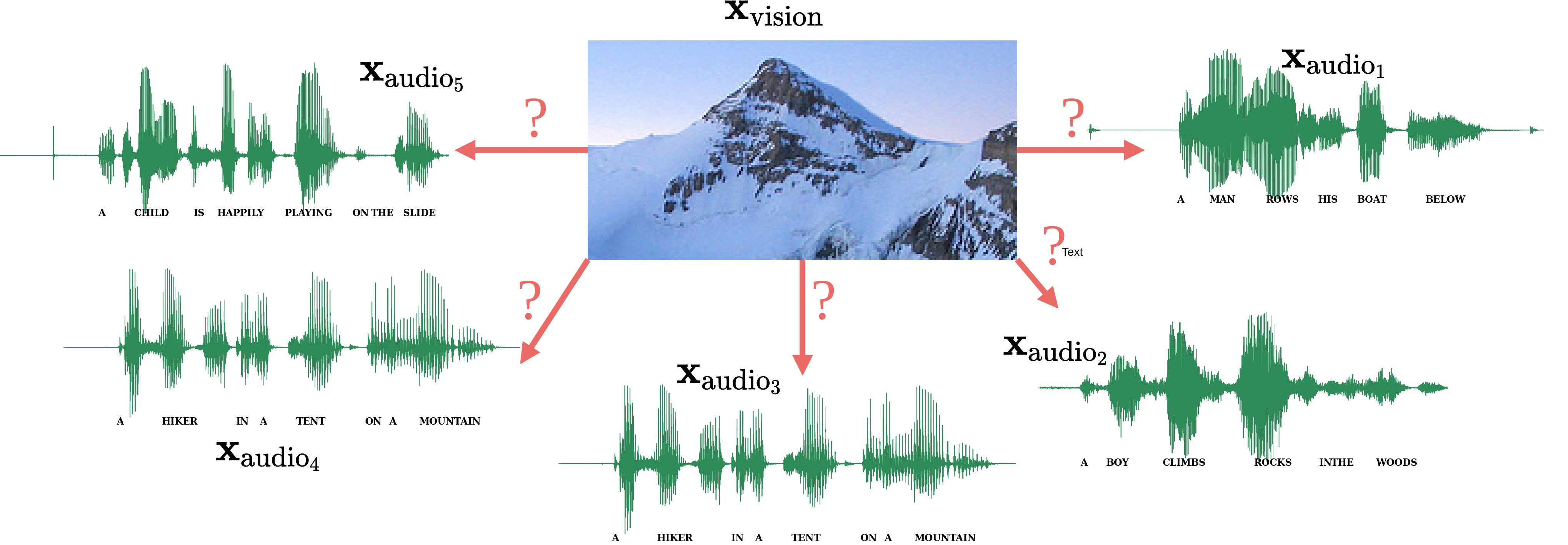}}
		\vspace{-12pt}
	\end{minipage}
	\caption{The goal in visually prompted detection is to detect whether a given query keyword (given as an image) occurs anywhere within a spoken utterance.}
	\label{fig:vpkd}
\end{figure}

The approach of directly training on image-speech pairs is motivated by children having access to image and speech signals when acquiring their native language~\cite{bomba_nature_1983, pinker_language_1994, eimas_studies_1994, roy_grounded_2003, boves_acorns_2007, gelderloos_phonemes_2016, rasanen_computational_2019}.
To learn words, they can use the co-occurrences of spoken words with
visual objects, and vice versa~\cite{miller_how_1987}.
Eventually humans can establish if and where a word depicted by
its visual representation,
 is uttered---without {ever} requiring 
 transcriptions.
Drawing inspiration from humans, we introduce the new task of \acf{vpkl}. 
This task is very similar to the task of textual keyword detection, where a model is given a written query keyword and asked to detect (and possibly locate) occurrences of the keyword in a search collection~\cite{garcia_keyword_2006, szoke_comparison_2005, wilpon_automatic_1990, kamper_visually_2017, kamper_visually_2018, olaleye_towards_2020, olaleye_attention-based_2021, olaleye_keyword_2022}.
Instead of a written keyword, in \ac{vpkl} the query is an image of an object or concept.

Formally, \ac{vpkl} involves both detection and localisation. 
Detection is illustrated in Fig.~\ref{fig:vpkd}.
A model is given an image query $\boldsymbol{\mathrm{x}}_{\textrm{vision}}$, which depicts a keyword, and asked whether the keyword occurs in an utterance $\boldsymbol{\mathrm{x}}_{\textrm{audio}}$.
For localisation, if the model detects the image query $\boldsymbol{\mathrm{x}}_{\textrm{vision}}$ in $\boldsymbol{\mathrm{x}}_{\textrm{audio}}$, the model is prompted to identify where in $\boldsymbol{\mathrm{x}}_{\textrm{audio}}$ the keyword occurs.
E.g.\ in Fig.~\ref{fig:vpkl}\ the model is asked whether the \textit{mountain} in $\boldsymbol{\mathrm{x}}_{\textrm{vision}}$, occurs in $\boldsymbol{\mathrm{x}}_{\textrm{audio}_3}$.
After the model detected the keyword ``mountain'', it is prompted to identify where in $\boldsymbol{\mathrm{x}}_{\textrm{audio}_3}$ it occurs.
During \ac{vpkl}, the model therefore has to first do detection and then localisation, i.e.\ detection is a task on its own, but localisation includes detection. 
To do \ac{vpkl}, we need a multimodal model that can output whether a keyword occurs and at which frame detected keywords occur. 

%% file: multimodalModel.tex
\section{Approach: Multimodal localisation models}
\label{sec:model} 

Our \ac{vpkl} model outputs an overall score $\mathcal{S}\in[0, 1]$ indicating whether a keyword is present anywhere within an utterance. The keyword is detected if the $\mathcal{S}$ is above a threshold $\alpha$.
Additionally, the model outputs a sequence of scores $\boldsymbol{\mathrm{a}}_{\textrm{audio}} \in \mathbb{R}^N$ where $N$ is the number of speech frames; each element $\mathrm{a}_{\textrm{audio}_i}$ indicates whether the detected keyword occurs at frame $i$.
To do this, we need a multimodal model that can predict which frames in an utterance is most relevant to a given query image.
As starting point for our model, we use the
\acf{davenet} of~\cite{harwath_unsupervised_2016}.
We then adapt it by introducing a new sampling scheme and attention mechanism that encourages localisation.

\subsection{Starting point: \ac{davenet}}
\label{subsec:baseline}

\ac{davenet}~\cite{harwath_unsupervised_2016} consists of a vision and an audio network which separately maps an image and its entire spoken caption to single fixed-size embeddings in a common multimodal space.
The goal is to get embeddings of paired images and spoken captions to be more similar than the embeddings of mismatched images and captions.
Our implementation of \ac{davenet} incorporates some of the extensions from \cite{harwath_jointly_2018}. 

Following \cite{harwath_jointly_2018}, we extend the \ac{davenet} architecture to use ResNet50~\cite{he_deep_2016} for the image network and instead of learning fixed-size embeddings, we learn a sequence of embeddings for each image $\boldsymbol{\mathrm{e}}_{\textrm{vision}} \in \mathbb{R}^M$ and caption $\boldsymbol{\mathrm{e}}_{\textrm{audio}} \in \mathbb{R}^N$.
Here $M$ is the number of pixels and $N$ the number of frames. 
These embedding sequences are then used in a \textit{matchmap} $\mathcal{M} \in \mathbb{R}^{M \times N}$ which calculates the dot product between each frame embedding in $\boldsymbol{\mathrm{e}}_{\textrm{audio}}$ and each pixel embedding in $\boldsymbol{\mathrm{e}}_{\textrm{vision}}$.
The idea is that high similarity in the 
$\mathcal{M}$ should indicate those speech frames and image pixels that are related.
In~\cite{harwath_jointly_2018}, the authors showed quantitatively that the matchmaps can indeed 
localise words and objects corresponding to the same concept.

Another change we make from \cite{harwath_unsupervised_2016} and \cite{harwath_jointly_2018}
is that, instead of using standard speech features as input, we use an acoustic network trained on external data. Concretely, we use a different network as the audio branch in our modified version of \ac{davenet}.
This network consists of an acoustic $f_{\textrm{acoustic}}$ and a BiLSTM $f_{\textrm{BiLSTM}}$ network. 
For $f_{\textrm{acoustic}}$, we pretrain the \ac{CPC} model of \cite{van_niekerk_vector-quantized_2020} on out-of-domain unlabelled data to
obtain more robust acoustic features (by taking advantage of more extensive data sources).
The acoustic features are sent to $f_{\textrm{BiLSTM}}$ 
to take advantage of 
context from 
the entire caption.
These changes improves the original \ac{davenet} model's caption retrieval score from $31\%$ to $69\%$ on the English Places corpus~\cite{harwath_unsupervised_2016, harwath_learning_2017, harwath_jointly_2018}.
We still refer to this custom implementation as \customDN, since it largely follows the original architecture.

We will see in Section~\ref{sec:experiments} that this multimodal model performs poorly on \ac{vpkl}. 
Therefore, we propose two improvements: a keyword sampling scheme (Section~\ref{subsec:sampling}) and a new localising attention mechanism (Section~\ref{subsec:multimodalAttention}). 

\subsection{Positive and negative keyword sampling}
\label{subsec:sampling}

To learn when an image query 
occurs in an utterance, we propose a keyword sampling scheme to push image-caption pairs containing the same 
keyword closer together and pairs not containing the keyword away from each other.
We could use an off-the-shelve visual tagger to generate textual tags for the images and use these predicted text labels to sample positive and negative image-caption pairs: if a keyword occurs in the predicted labels of two image-caption pairs, they are positives; if not, they are negatives. 

We mainly consider this approach with a perfect tagger, but later in \S~\ref{subsec:further_analysis} we do initial experiments using a real tagger.
In 
{the} idealised case we use the ground truth textual transcriptions of the spoken captions to sample positive $(\boldsymbol{\mathrm{x}}_{\textrm{audio}}^+,\, \boldsymbol{\mathrm{x}}_{\textrm{vision}}^+)$ and three negative $(\boldsymbol{\mathrm{x}}_{\textrm{audio}_i}^-,\, \boldsymbol{\mathrm{x}}_{\textrm{vision}_i}^-)$ pairs for each image-caption $(\boldsymbol{\mathrm{x}}_{\textrm{audio}},\, \boldsymbol{\mathrm{x}}_{\textrm{vision}})$ pair.
%
E.g.\ given an image-caption pair $(\boldsymbol{\mathrm{x}}_{\textrm{audio}},\, \boldsymbol{\mathrm{x}}_{\textrm{vision}})$ where $\boldsymbol{\mathrm{x}}_{\textrm{audio}}$ is ``a hiker in a tent on a mountain'', we choose a positive pair also containing the keyword ``mountain'' like  ``a mountain with a snowy peak''. 
Negatives are selected from images that do not contain the keyword ``mountain'' in the transcription of their captions.

Since we use these pairs in a contrastive loss, we refer to this model as \contrastiveDN. 
The model is trained to push the similarity scores $\mathcal{S}$ of pairs containing a certain keyword closer and the similarity scores $\mathcal{S}$ of pairs not containing the keyword further apart:
\begin{equation*}\scriptsize\begin{split}
		& l(\boldsymbol{\mathrm{e}}_{\textrm{audio}}, \boldsymbol{\mathrm{e}}_{\textrm{vision}}, \boldsymbol{\mathrm{e}}_{\textrm{audio}}^+, \boldsymbol{\mathrm{e}}_{\textrm{vision}}^+, \boldsymbol{\mathrm{e}}_{\textrm{audio}_{1:3}}^-, \boldsymbol{\mathrm{e}}_{\textrm{vision}_{1:3}}^-) \\
		&= \,2\times\textrm{MSE}\Big(\mathcal{S}(\boldsymbol{\mathrm{e}}_{\textrm{vision}}, \boldsymbol{\mathrm{e}}_{\textrm{audio}}), 1\Big)\\
		&+ \,\textrm{MSE}\Big(\mathcal{S}(\boldsymbol{\mathrm{e}}_{\textrm{vision}}, \boldsymbol{\mathrm{e}}_{\textrm{audio}}^+), 1\Big)  +  \textrm{MSE}\Big(\mathcal{S}(\boldsymbol{\mathrm{e}}_{\textrm{audio}}, \boldsymbol{\mathrm{e}}_{\textrm{vision}}^+), 1\Big) \\
		&+ \,\sum_{i=1}^{3}\Bigg( \textrm{MSE}\Big(\mathcal{S}(\boldsymbol{\mathrm{e}}_{\textrm{vision}}, \boldsymbol{\mathrm{e}}^-_{\textrm{audio}_i}), -1\Big) + \textrm{MSE}\Big(\mathcal{S}(\boldsymbol{\mathrm{e}}_{\textrm{audio}}, \boldsymbol{\mathrm{e}}^-_{\textrm{vision}_i}). -1\Big)\Bigg),\end{split}\end{equation*}
We obtain $\mathcal{S}$ by taking the matchmap $\mathcal{M}$ between the given embedding sequences before max-pooling over the pixel axis and mean-pooling over the temporal axis, as in~\cite{harwath_jointly_2018}.
We use the mean-squared-error (MSE) 
between the embeddings 
with a positive target of $1$ and a negative target of $-1$.
This was based on 
development experiments: we found that MSE and the wider range of possible target values gave the best results since it pushed the positives and negatives further apart.  

Note 	that \contrastiveDN\ is the step between \customDN\ and our proposed attention model,  
described next.

\subsection{Localising attention mechanism}
\label{subsec:multimodalAttention}

\begin{figure}[!tb]
	
	\begin{minipage}[b]{1.0\linewidth}
		\centering
		\centerline{\includegraphics[width=\linewidth]{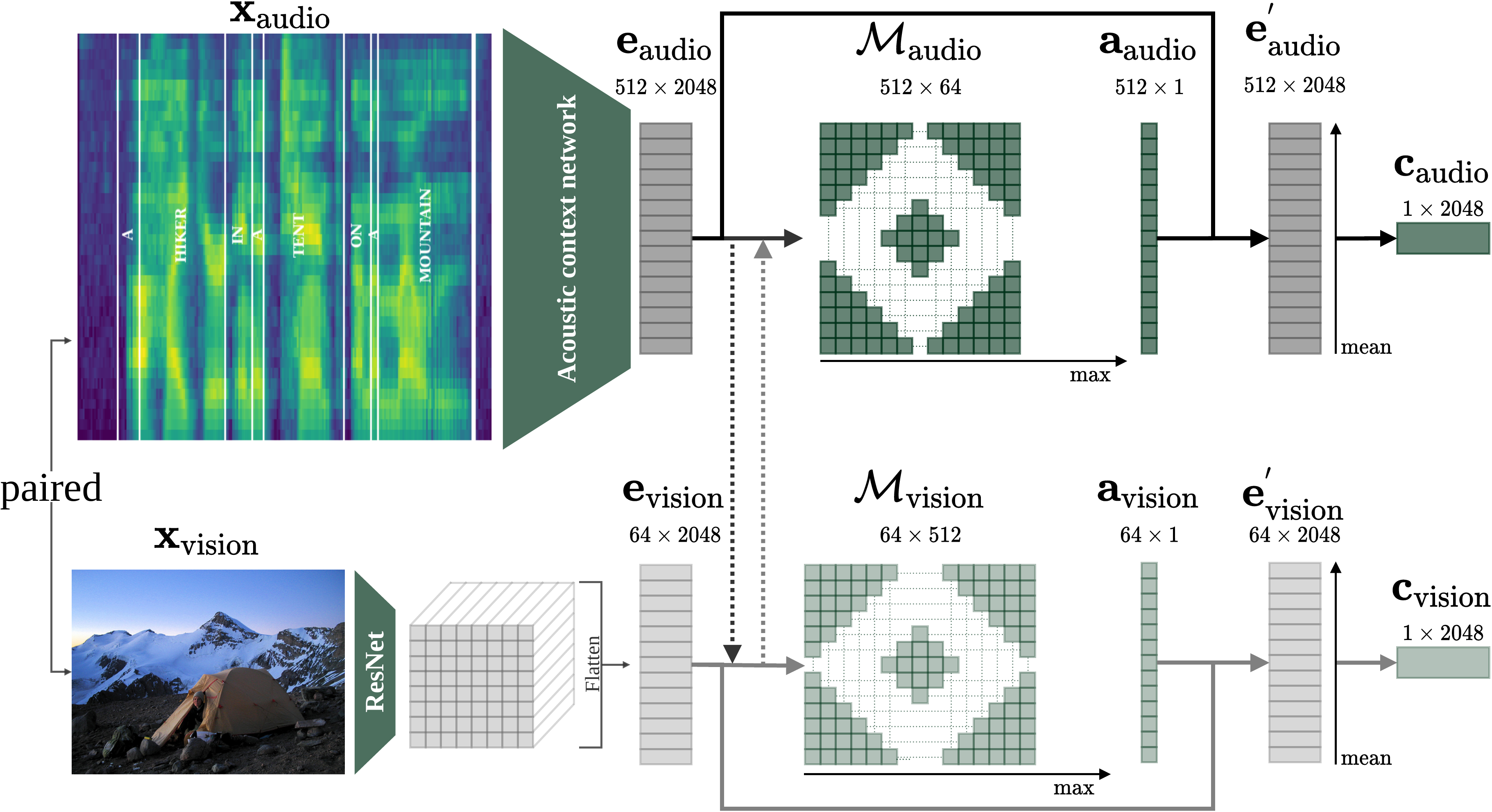}}
		\vspace{-12pt}
	\end{minipage}
	\caption{The \ma\ takes a pair $(\boldsymbol{\mathrm{x}}_{\textrm{audio}},\, \boldsymbol{\mathrm{x}}_{\textrm{vision}})$ as input. The pair given to the model can also be a positive pair $(\boldsymbol{\mathrm{x}}_{\textrm{audio}}^+,\, \boldsymbol{\mathrm{x}}_{\textrm{vision}})$ or $(\boldsymbol{\mathrm{x}}_{\textrm{audio}},\, \boldsymbol{\mathrm{x}}_{\textrm{vision}}^+)$, or a negative pair $(\boldsymbol{\mathrm{x}}_{\textrm{audio}},\, \boldsymbol{\mathrm{x}}_{\textrm{vision}_i}^-)$ or $(\boldsymbol{\mathrm{x}}_{\textrm{audio}_i}^-,\, \boldsymbol{\mathrm{x}}_{\textrm{vision}})$.}
	\label{fig:multimodal_attention}
	\vspace{-5pt}
\end{figure}

We make use of an attention mechanism similarly to \cite{olaleye_attention-based_2021, olaleye_keyword_2022, palaz_jointly_2016, tamer_keyword_2020}. 
These models all take a written keyword as a query, and then look up a word embedding corresponding to that keyword, which is then used to calculate per-frame attention weights $\boldsymbol{\mathrm{a}}_{\textrm{audio}}$ over the per-frame audio embeddings $\boldsymbol{\mathrm{e}}_{\textrm{audio}}$.
In our case we instead use the matchmap $\mathcal{M}_{\textrm{audio}}$, as shown in Fig.~\ref{fig:multimodal_attention}.
As a reminder, the audio matchmap $\mathcal{M}_{\textrm{audio}}$ is calculated by taking the dot product between each audio embedding in $\boldsymbol{\mathrm{e}}_{\textrm{audio}}$ and each vision embedding in $\boldsymbol{\mathrm{e}}_{\textrm{vision}}$.
We then take the maximum over $\mathcal{M}_{\textrm{audio}}$'s pixel axis to obtain $\boldsymbol{\mathrm{a}}_{\textrm{audio}}$.
Thereafter, $\boldsymbol{\mathrm{a}}_{\textrm{audio}}$ is used to weigh $\boldsymbol{\mathrm{e}}_{\textrm{audio}}$ to get $\boldsymbol{\mathrm{e}}^{'}_{\textrm{audio}}$.
From $\boldsymbol{\mathrm{e}}^{'}_{\textrm{audio}}$, we sum over the temporal axis to get a context vector $\boldsymbol{\mathrm{c}}_{\textrm{audio}}$.
In the vision branch we do something very similar: we take the maximum over $\mathcal{M}_{\textrm{vision}}$'s temporal axis to obtain $\boldsymbol{\mathrm{a}}_{\textrm{vision}}$ from which we can calculate $\boldsymbol{\mathrm{e}}^{'}_{\textrm{vision}}$ and $\boldsymbol{\mathrm{c}}_{\textrm{vision}}$ (similar to how we obtained $\boldsymbol{\mathrm{e}}^{'}_{\textrm{audio}}$ and $\boldsymbol{\mathrm{c}}_{\textrm{audio}}$).
It is important to note that $\mathcal{M}_{\textrm{vision}}$ is the transpose of $\mathcal{M}_{\textrm{audio}}$.

We train this model, which we refer to as \ma, by using the $\boldsymbol{\mathrm{c}}$ vectors in a contrastive loss:
\begin{equation*}\scriptsize\begin{split}
	  &l(\boldsymbol{\mathrm{c}}_{\textrm{audio}}, \boldsymbol{\mathrm{c}}_{\textrm{vision}}, \boldsymbol{\mathrm{c}}_{\textrm{audio}}^+, \boldsymbol{\mathrm{c}}_{\textrm{vision}}^+, \boldsymbol{\mathrm{c}}_{\textrm{audio}_{1:3}}^-, \boldsymbol{\mathrm{c}}_{\textrm{vision}_{1:3}}^-) \\
	  &= \sum_{i=1}^{3} \textrm{MSE}\Big(\mathcal{S}(\boldsymbol{\mathrm{c}}_{\textrm{audio}}, \boldsymbol{\mathrm{c}}_{\textrm{vision}}), 1\Big) + \textrm{MSE}\Big(\mathcal{S}(\boldsymbol{\mathrm{c}}_{\textrm{audio}}, \boldsymbol{\mathrm{c}}_{\textrm{audio}}^+), 1\Big) \\
	  &+ \textrm{MSE}\Big(\mathcal{S}(\boldsymbol{\mathrm{c}}_{\textrm{audio}}, \boldsymbol{\mathrm{c}}^-_{\textrm{audio}_i}), -1\Big) + \textrm{MSE}\Big(\mathcal{S}(\boldsymbol{\mathrm{c}}_{\textrm{vision}}, \boldsymbol{\mathrm{c}}_{\textrm{audio}}), 1\Big)\\
	  &+ \textrm{MSE}\Big(\mathcal{S}(\boldsymbol{\mathrm{c}}_{\textrm{vision}}, \boldsymbol{\mathrm{c}}_{\textrm{audio}}^+), 1\Big) + \textrm{MSE}\Big(\mathcal{S}(\boldsymbol{\mathrm{c}}_{\textrm{vision}}, \boldsymbol{\mathrm{c}}^-_{\textrm{vision}_i}), -1\Big),\end{split}
\end{equation*}
where $\mathcal{S}$ is the cosine similarity. 
This loss attempts to push $\boldsymbol{\mathrm{c}}_{\textrm{audio}}$ closer to its paired image $\boldsymbol{\mathrm{c}}_{\textrm{vision}}$ and to a caption also containing the same keyword $\boldsymbol{\mathrm{c}}_{\textrm{audio}}^+$.
At the same time, it should push $\boldsymbol{\mathrm{c}}_{\textrm{audio}}$ away from any caption $\boldsymbol{\mathrm{c}}_{\textrm{audio}_i}^-$ not containing the keyword.
The same goes for the vision part.
The idea is to force the model to isolate the keywords common to image-caption pairs and their positive samples. 


%% file: experimentalSetup.tex
\section{Experiments}
\label{sec:experiments}

The goal in \ac{vpkl} is to detect and localise a visual keyword in a spoken utterance.
To do this, we train three speech-vision models to predict which (if any) of the frames are related to a given image query.

\subsection{Experimental setup}
\label{subsec:setup}

\textbf{Data:} We pretrain the acoustic network $f_{\textrm{acoustic}}$ on the combination of LibriLight~\cite{kahn_libri-light_2020} and the multilingual Places dataset~\cite{harwath_vision_2018}.
We train \customDN, \contrastiveDN\ and \ma\ on the Flickr8k Audio Captions Corpus~\cite{harwath_deep_2015}.
The corpus consists of 8k images where each image is paired with five parallel spoken English captions and is split into train, development and test sets of 30k, 5k and 5k utterances, respectively.
The spoken  captions are parametrised as mel-spectograms, with a hop length of 10~ms, a window width of 25~ms and 40~mel bins.
We truncate or zero-pad the mel-spectograms to 1024 frames. 
Images are resized to 224$\times$224 pixels {and} 
normalised according to the VGG~\cite{simonyan_very_2015} mean and variances calculated using~ImageNet~\cite{deng_imagenet_9}.

\textbf{Models:} For the image branch, we use the ResNet50~\cite{he_deep_2016} adaptation of \cite{harwath_learning_2020}.
The architecture used for the audio branch of all three models is given in the source code.\footnote{\smallcustomDN: \url{https://github.com/LeanneNortje/DAVEnet_VPKL}; \\\smallcontrastiveDN: \url{https://github.com/LeanneNortje/ContrastiveDAVEnet_VPKL}; \\\smallma: \url{https://github.com/LeanneNortje/LocalisationAttentionNet_VKPL}}
For training \contrastiveDN\ and \ma\ we use the training set's transcriptions to sample positive pairs $(\boldsymbol{\mathrm{x}}_{\textrm{audio}}^+,\, \boldsymbol{\mathrm{x}}_{\textrm{vision}}^+)$ and three negative pairs $(\boldsymbol{\mathrm{x}}_{\textrm{audio}_i}^-,\, \boldsymbol{\mathrm{x}}_{\textrm{vision}_i}^-)$ 
for each image-caption $(\boldsymbol{\mathrm{x}}_{\textrm{audio}},\, \boldsymbol{\mathrm{x}}_{\textrm{vision}})$ pair.
{For validation on \contrastiveDN\ and \ma, we follow the same procedure to sample a positive pair $(\boldsymbol{\mathrm{x}}_{\textrm{audio}}^+,\, \boldsymbol{\mathrm{x}}_{\textrm{vision}}^+)$ and a negative pair $(\boldsymbol{\mathrm{x}}_{\textrm{audio}}^-,\, \boldsymbol{\mathrm{x}}_{\textrm{vision}}^-)$ for each image-caption $(\boldsymbol{\mathrm{x}}_{\textrm{audio}},\, \boldsymbol{\mathrm{x}}_{\textrm{vision}})$ pair using the development set. 
The validation task measures whether the model will place the positive pair closer to  $(\boldsymbol{\mathrm{x}}_{\textrm{audio}},\, \boldsymbol{\mathrm{x}}_{\textrm{vision}})$ than it would the negative pair. 
\customDN\ is trained and validated on Flickr8k following a similar setup to the original paper~\cite{harwath_unsupervised_2016}.}
All models are trained with Adam~\cite{kingma_adam_2015} for 100 epochs using early stopping.
The batch sizes and learning rates for each model are tuned on development data. 

\textbf{Evaluation:} We evaluate our approach on a set of 34 keywords. This is a subset of the keywords from~\cite{kamper_semantic_2019}, where we only consider keywords that can reasonably be localised using an image.
For each keyword, we obtain 10 images from the Flickr8k test split and then manually crop the image region corresponding to the keyword.\footnote{\ac{vpkl} task: \url{https://github.com/LeanneNortje/VPKL}}
This cropped image then serves as the image query for that keyword.
At test time, we obtain $\boldsymbol{\mathrm{c}}_{\textrm{audio}}$ for a test utterance and $\boldsymbol{\mathrm{c}}_{\textrm{vision}}$ for an image query.
These are compared using 
cosine similarity.
If the 
score is above a threshold $\alpha$ for a given caption and image query, the keyword is detected in the caption.
The $\alpha$ for each model is tuned on the development set: 
$\alpha_{\textrm{\smallcustomDN}} = 0.85$, $\alpha_{\textrm{\smallcontrastiveDN}} = 0.55$ and $\alpha_{\textrm{\smallma}} = 0.5885$.
For localisation, if a keyword is detected, then the position of the keyword is taken as the frame where the maximum attention weight occurs. 
This is compared to ground truth alignments, with a true positive occurring when the predicted frame
falls within the time-span of the keyword in the reference. 
If a keyword is not detected or the prediction does not fall within the time-span of the keyword, this is counted as a mistake.
{We also implement a random baseline, where a detection score is randomly sampled between $-1$ and $1$ and the attention weights for each frame is randomly assigned a value between $0$ and $1$.}

\textbf{Visual \ac{bow} baseline:} 
Previous work~\cite{kamper_visually_2017, kamper_visually_2018} used a visually grounded \ac{bow} model to detect written keywords.
They use a visual tagger to extract \ac{bow} labels for each training image, which then serves as targets for the corresponding spoken caption.
The resulting model can be used to detect written keywords in utterances.
This was extended to not just detect whether the keyword occurs in the utterance, but also to determine where in the utterance the word occurs (if it was detected)~\cite{olaleye_towards_2020, olaleye_attention-based_2021, olaleye_keyword_2022}.
We compare our models to their model, which takes a written query rather than an image query as input.
The visual \ac{bow} model is evaluated on the same 34 keywords as our visually prompted models.

%% file: results.tex
\subsection{Results}
\label{subsec:results}

\begin{table}[!b]
	\vspace{-15pt}	
	\renewcommand{\arraystretch}{1.1}
	\centering
	\setlength\tabcolsep{3pt}
	\caption{Keyword detection results (\%).}
	\small
	\begin{tabularx}{\linewidth}{@{}Lccc@{}}
		\toprule
		Model & Precision & Recall & F1 score\\
		\midrule 
		\textit{Text query} & & & \\
		Visually grounded \ac{bow}~\cite{olaleye_towards_2020} & 42.29 & 36.32 & 39.08 \\
		\addlinespace
		\textit{Image query} & & & \\
		Random baseline & 2.30 & 13.96 & 3.94 \\
		\smallcustomDN & 8.86 & 46.51 & 14.88 \\
		\smallcontrastiveDN & 37.97 & 44.84 & 41.12 \\
		\smallma & \textbf{48.41} & \textbf{55.85} & \textbf{51.86} \\
		\bottomrule
	\end{tabularx}
	\label{tbl:detection}
\end{table}

\begin{table}[!t]
	\renewcommand{\arraystretch}{1.1}
	\centering
	\setlength\tabcolsep{3pt}
	\caption{Keyword localisation results (\%).}
	\small
	\begin{tabularx}{\linewidth}{@{}Lccc@{}}
		\toprule
		Model & Precision & Recall & F1 score\\
		\midrule 
		\textit{Text query} & & & \\
		Visually grounded \ac{bow}~\cite{olaleye_towards_2020} & 33.39 & 31.02 & 32.17 \\
		\addlinespace
		\textit{Image query}  & & & \\
		Random baseline & 0.13 & 0.87 & 0.22 \\
		\smallcustomDN & 5.17 & 33.36 & 8.95 \\
		\smallcontrastiveDN & 30.43 & 39.45 & 34.36 \\
		\smallma & \textbf{44.43} & \textbf{53.77} & \textbf{48.66} \\
		\bottomrule
	\end{tabularx}
\label{tbl:localisation}
\vspace{-10pt}
\end{table}

\begin{figure*}[!t]
	
	\begin{minipage}[b]{1.0\linewidth}
		\centering
		\centerline{\includegraphics[width=0.99\linewidth]{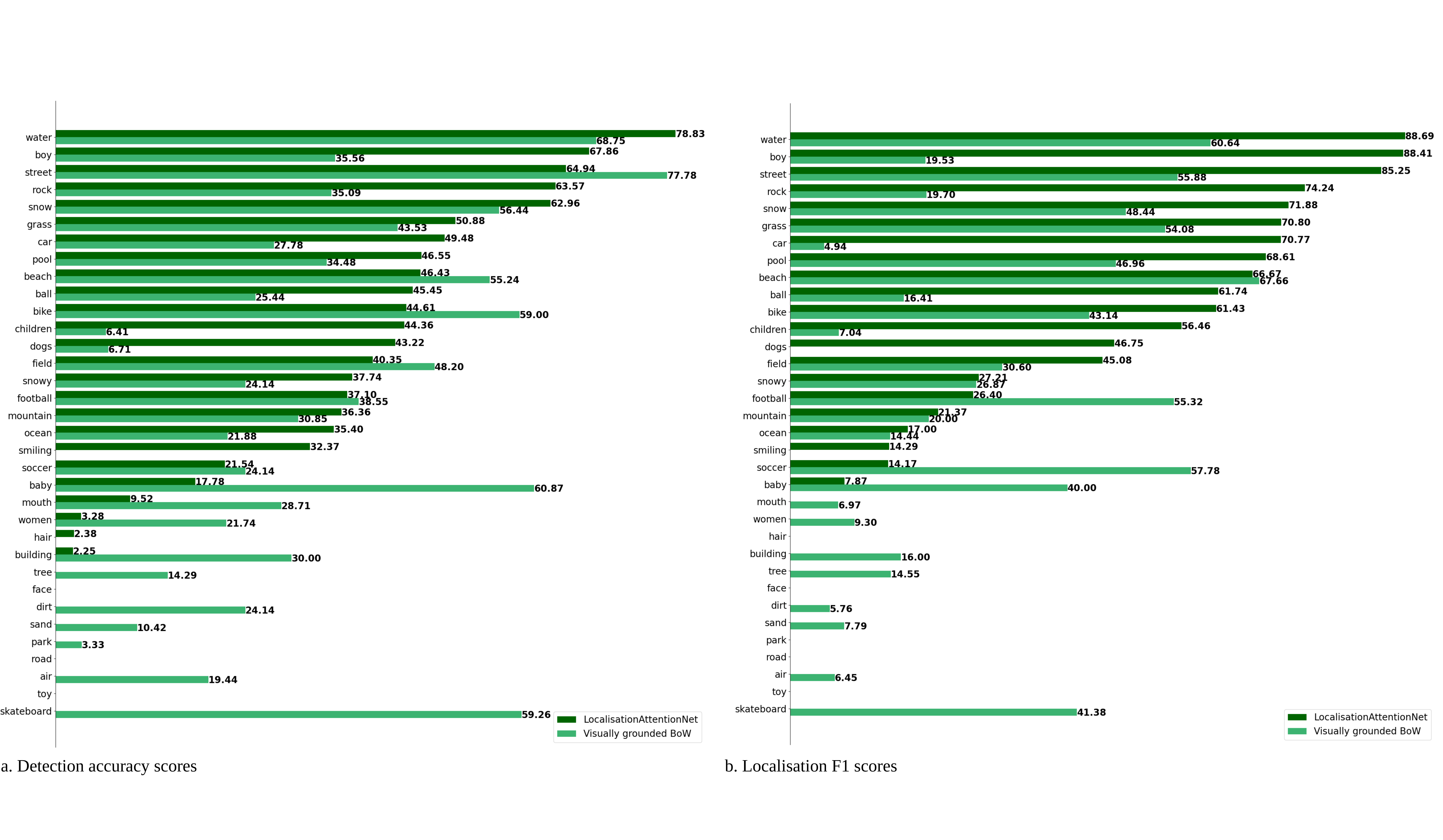}}
		\vspace{-25pt}
	\end{minipage}
	\caption{The per keyword (a) detection accuracy scores and (b) localisation F1 score achieved by \ma\ 
		and the visually grounded \ac{bow} model.} 
	\label{fig:det_acc_and_loc_f1}
	\vspace{-10pt}
\end{figure*}

Tables~\ref{tbl:detection} and~\ref{tbl:localisation} compare our \ac{vpkl} models to the visual \ac{bow} model for keyword detection and localisation.
We see that \customDN\ outperforms the random baseline but not the visually grounded \ac{bow} model.
Moreover, \contrastiveDN\ outperforms this \ac{bow} model on recall and F1, but not in precision. 
The improvements of \contrastiveDN\ over \customDN\ shows that by sampling positives and negatives using the scheme introduced in Section~\ref{subsec:sampling}, we can better detect and localise visual keywords.

Building on this, \ma\ improves not only the precision and F1 scores of \contrastiveDN, but also the recall. 
\ma\ outperforms the visually grounded \ac{bow} model which shows that \ac{vpkl} can be more accurate than keyword embedding localisation if an appropriate speech-vision model with a localisation objective is used.  

\subsection{Further analyses}
\label{subsec:further_analysis}

\begin{figure}[!t]
    
    \begin{minipage}[b]{0.99\linewidth}
        \centering
        \centerline{\includegraphics[width=\linewidth]{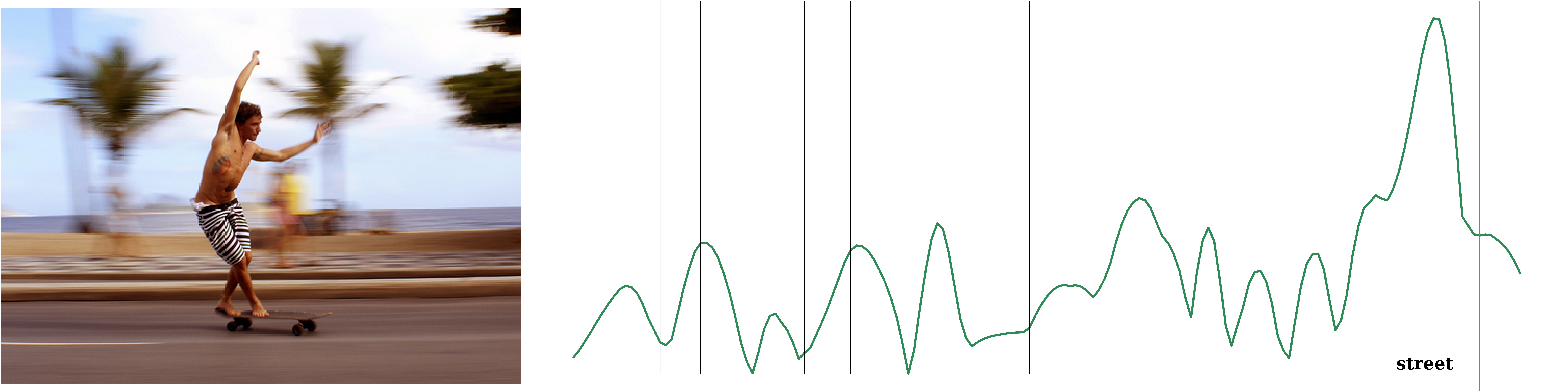}}
        
        \centerline{\includegraphics[width=\linewidth]{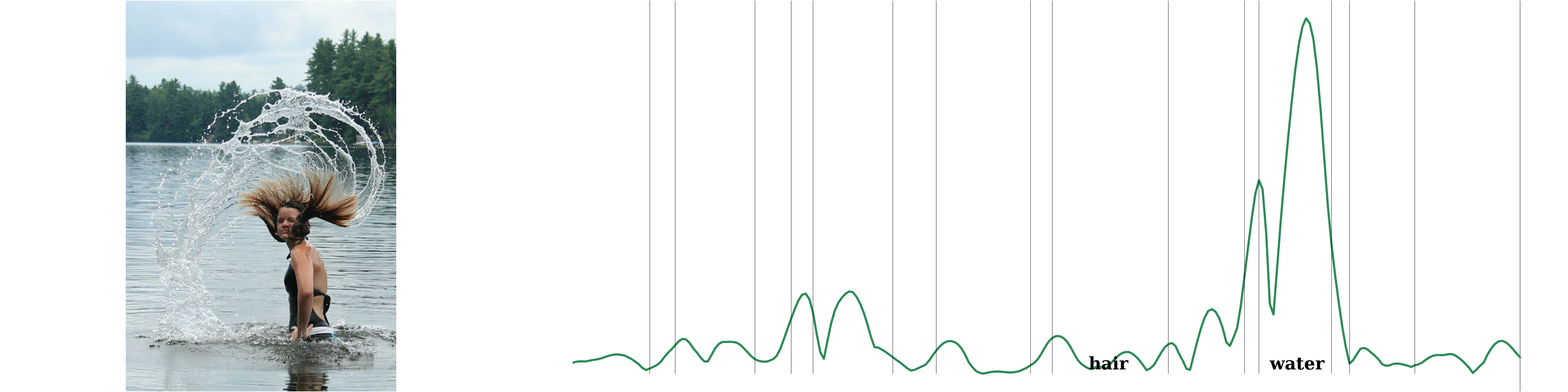}}
        \vspace{-10pt}
    \end{minipage}
    \caption{The audio attention weights from \ma\ for two utterance-query pairs.}
\label{fig:loc_example}
\vspace{-10pt}
\end{figure}

The metrics above
are based on aggregating scores across keyword types; we now consider per-keyword performance to get a better understanding of how and when the \ac{vpkl} models outperform the visual \ac{bow} model.
Fig.~\ref{fig:det_acc_and_loc_f1} shows the (a) detection accuracy scores and (b) localisation F1 scores for each keyword for \ma\ and the visual \ac{bow} model.
We see that \ma\ learns to detect and localise certain keywords more accurately than the \ac{bow} model. 
However, \ma\ detects and localises a smaller distribution of keywords, especially when it comes to localisation.
E.g., we see worse performance for keywords corresponding to verbs, which makes sense since these are harder to depict with images: ``sits'', ``sitting'', ``riding'' and ``rides''.
Other keywords that the model struggles with are colours (e.g.\ ``orange'') or keywords that are very general (e.g.\ ``air'').

We also see that \ma\ can detect some of the keywords but not localise them accurately.
In order to see why this happens, we qualitatively look at attention weights produced by the model when given an image query. 
Fig.~\ref{fig:loc_example} shows two examples. 
In the top example, only one keyword ``street'' is present, and in the bottom example two keywords ``water'' and ``hair'' are present in the query.
The model can output multiple detection scores for an image query but only one set of attention scores. 
I.e.\ we might be able to detect both ``water'' and ``hair'', but we can only localise one keyword. 
This problem cannot be fixed by taking a smaller crop around \textit{hair} for the keyword ``hair'', since the image would still visually contain \textit{water}.
Therefore, the model cannot localise two keywords when both are present in an image query.

\begin{table}[!t]
	\renewcommand{\arraystretch}{1.1}
	\centering
	\setlength\tabcolsep{3pt}
	\caption{Keyword detection results (\%) on the initial 34 keywords.}
	\small
	\begin{tabularx}{\linewidth}{@{}Lccc@{}}
		\toprule
		Model & Precision & Recall & F1 score\\
		\midrule 
		Random baseline & 2.30 & 13.96 & 3.94 \\
		\smallcustomDN & 8.86 & 46.51 & 14.88 \\
		\smallma & \textbf{48.41} & \textbf{55.85} & \textbf{51.86} \\
		\smallma\ trained on generated tags for 34 keywords & 31.02 & 31.83 & 31.42 \\
		\smallma\ trained on 190 ground truth keywords & 20.77 & 37.06 & 26.62 \\
		\bottomrule
	\end{tabularx}
	\label{tbl:detection_ov}
	\vspace{-10pt}
\end{table}

\begin{table}[!t]
	\renewcommand{\arraystretch}{1.1}
	\centering
	\setlength\tabcolsep{3pt}
	\caption{Keyword localisation results (\%) on the initial 34 keywords.}
	\small
	\begin{tabularx}{\linewidth}{@{}Lccc@{}}
		\toprule
		Model & Precision & Recall & F1 score\\
		\midrule 
		Random baseline & 0.13 & 0.87 & 0.22 \\
		\smallcustomDN & 5.17 & 33.36 & 8.95 \\
		\smallma & \textbf{44.43} & \textbf{53.77} & \textbf{48.66} \\
		\smallma\ trained on generated tags for 34 keywords & 23.20 & 25.75 & 24.21 \\
		\smallma\ trained on 190 ground truth keywords & 14.48 & 28.93 & 19.30 \\
		\bottomrule
	\end{tabularx}
	
	\label{tbl:localisation_ov}
	\vspace{-10pt}
\end{table}

To investigate the scalability of \smallma, we extend the number of keywords to be learnt to 190.\footnote{\smallma\ on 190 ground truth keywords: \url{https://github.com/LeanneNortje/LocalisationAttentionNet_VPKL_on_190_keywords}}
These include the initial 34 keywords.
The results in Table~\ref{tbl:detection_ov} and \ref{tbl:localisation_ov}  (in line 5) show that extending the number of keywords leads to a decrease in performance on the original 34 keywords.
However, our approach still outperforms \customDN, which has no form of keyword learning. 

In moving towards open vocabulary \ac{vpkl} in zero-resource settings, it is therefore clear that some form of keyword learning would be beneficial.
\ac{vpkl} with an open vocabulary would ideally allow for detection of any keyword depicted using a visual image query. 
An evaluation on more keywords is one step in this direction, but we are still considering an idealised case using transcriptions to find positive and negative examples. 
We therefore also present initial experiments where we instead use an actual visual tagger.\footnote{\smallma\ on generated tags: \url{https://github.com/LeanneNortje/LocalisationAttentionNet_VPKL_on_generated_tags}}



{For this approach on \smallma, we use the of-the-shelf visual tagger of \cite{kamper_semantic_2019} to tag the training images with possible keywords.
To sample positive and negative image-caption pairs, we then use these predicted keywords. 
We only use the pairs in which one or more of the 34 keywords (\S\ref{subsec:setup}) occurs to sample positive and negatives: pairs with the same predicted keywords are positives and pairs with different predicted keywords are negatives.
From Table~\ref{tbl:detection_ov} and \ref{tbl:localisation_ov}, we see that this approach (line 4) outperforms \customDN\ (line 2). 
We conclude that any of our keyword sampling approaches, whether supervised or unsupervised, leads to keyword learning.}

{Both the generated keyword approach and the 190-word extended keyword evaluations fall short of \smallma\ trained on ground truth keywords. 
Therefore, the question remains: how can we extend this approach to work 
on zero-resource languages? 
Future work could look into finding a more accurate visual tagger.
However, the problem of the model only being able to learn a small amount of keywords remains.}
{We recommend looking into multimodal few-shot learning: learning a new keyword from a few paired images and spoken captions containing the keyword~\cite{eloff_multimodal_2019}.
Recently, \cite{miller_t_exploring_2022} looked into few-shot learning using natural images and spoken captions to learn new concepts, but they required a large number of examples to learn a new concept.
Although \cite{nortje_unsupervised_2020, nortje_direct_2021, eloff_multimodal_2019} only learned 
digit classes, they showed that new concepts can be learned for only a few examples.
The hope is that few-shot learning would enable \smallma\ 
to learn more keywords.} 

%% file: conclusions.tex
\section{Conclusion}
\label{sec:conclusions}

We proposed the new \ac{vpkl} task and adapted previous speech-vision models to perform better on this task.
Concretely, we proposed a new data sampling scheme, where we use images to find positive utterance pairs containing the same keyword, and we proposed a new localising attention mechanism over matchmaps.
We showed that both these innovations gave improvements in \ac{vpkl} over previous models.

In our main experiments
we used the transcriptions of the spoken captions in order to simulate an ideal tagger for our sampling scheme.
In further initial experiments,
 we also adapted the sampling approach to use a real tagger to generate image tags. 
Although the method using an ideal tagger outperformed the one using a real tagger, both methods outperformed \customDN, which uses no keywords sampling method. 
This proves that our sampling approach leads to keyword learning.
We also implemented a model with a larger vocabulary size using the ground truth transcriptions.
This caused a drop in performance compared to the case where less keywords are learned, but still gave improvements over  \customDN.
Future work will look into using few-shot learning to 
learn more keywords 
for zero-resource systems. 